\documentclass{article}

\usepackage{subfigure}
\usepackage{amsfonts}
\usepackage{amssymb}
\usepackage{pseudocode}
\usepackage{amsthm}
\usepackage{latexsym}
\usepackage{xspace}
\usepackage{graphicx}
\usepackage{amsmath}
\usepackage[latin1]{inputenc}
\usepackage{epsfig}
\usepackage{mathrsfs}

\newtheorem{ex}{Example}[section]

\newcommand{\bone}{\boldsymbol{1}}

\newcommand{\bx}{\boldsymbol{x}}
\newcommand{\bw}{\boldsymbol{w}}
\newcommand{\by}{\boldsymbol{y}}

\newcommand{\bzero}{\boldsymbol{0}}

\begin{document}
\title{Ensemble Models with Trees and Rules}
\author{Deniz Akdemir \\ Department of Plant Breeding \& Genetics\\ 
  Cornell University\\ Ithaca, NY}

\maketitle

\begin{abstract} 
In this article, we have proposed several approaches for post processing a large ensemble of regression trees or conjunctive rules.  These techniques are applied to  estimation of quantitative traits from markers, to the benchmark ''Boston Housing'' data set and to some simulated data. The results from these experiments show that the methods we have considered here are promising. In most cases, the models constructed by post processing the learners with partial least squares regression had better prediction performance than, for example, the ones produced by the random forest or the rulefit algorithms which use equal weights or weights estimated from lasso regression.     
\end{abstract}
{Keywords \& Phrases: Decision Trees, Ensemble Learning, Rules, Regression}
\section{Review of Ensemble Methods}
Ensemble learning (\cite{ho1990combination},  \cite{hansen1990neural},  \cite{kleinberg1990stochastic}) provides solutions to complex statistical prediction problems by simultaneously using a number of models. By bounding false idealizations, focusing on regularities and stable common behavior,  ensemble modeling approaches provide solutions that as a whole outperform the single models. 

Some early developments in ensemble learning include by Breiman with Bagging (bootstrap aggregating) (\cite{breiman1996bagging}) and random forest (\cite{breiman2001random}), and Freund and Shapire  with AdaBoost (\cite{freund1996experiments}). These methods involve ''random'' sampling the ''space of models'' to produce an ensemble of base learners and a ''post-processing'' of these to construct a final prediction model.  

In this article, we review several different approaches for ensemble post-processing and propose some new ones. The main point of this article is that the base learners in an ensemble can be used as an input to any regression model. The choices of different models here in are based on the experience and preferences of the authors.  

In the remainder of this section, we will review the recently proposed important sampling learning ensembles (ISLE) framework \cite{friedman2003importance} for ensemble model generation. The rule ensembles are also reviewed herein. In Section 2, we propose new ensemble post processing methods including partial least squares regression, multivariate kernel smoothing and use of out-of-bag observations. Section 3 is reserved for examples and simulations by which we compare the methods proposed here with the existing ones. Some remarks about hyper parameter choice and directions for future research are provided in Section 4.

\subsection{ISLE Approach}

Given a learning task and a relevant data set, we can generate a set of models from a predetermined model family.   Bagging bootstraps the training data set \cite{breiman1996bagging} and produces a model for each bootstrap sample. Random forest (\cite{ho1995random, breiman2001random}) creates a diverse set of  models by randomly selecting a few aspects of the data set while generating each model. AdaBoost \cite{freund1996experiments} and ARCing \cite{breiman1998arcing} iteratively build models by varying case weights (up-weighting cases with large current errors and down-weighting those accurately estimated) and employs the weighted sum of the estimates of the sequence of models. There have been few attempts to unify these ensemble learning methods. One such framework is the ISLE due to Popescu \& Friedman \cite{friedman2003importance}.

We are to produce a regression model to predict the continuous outcome variable $y$ from $p$ vector of input variables $\bx.$ We will generate models from a given model family $\mathscr{F}=\{f(\bx, \theta): \theta \in \Theta\}$ indexed by the parameter $\theta.$ The final ensemble models considered by  the ISLE framework  have an additive form: 
\begin{equation}F(\bx)=w_0+\sum_{j=1}^{M}w_{j} f(\bx, \theta_j)\label{eq:additivemodel}\end{equation}
where $\{f(\bx, \theta_j)\}_{j=1}^{M}$ are base learners selected from $\mathscr{F}.$ ISLE uses a two-step approach to produce $F(\bx)$. The first step involves sampling the space of possible models to obtain $\{\widehat{\theta}_j\}_{j=1}^{M}$.  The second step proceeds with combining the base learners by choosing weights $\{w_j\}_{j=0}^{M}$ in (\ref{eq:additivemodel}). 

The pseudo code to produce $M$ models $\{f(\bx, \widehat{\theta}_j)\}_{j=1}^{M}$ under ISLE framework is given below:

\begin{pseudocode}{ISLE}{M, \nu, \eta}
\label{ISLE}

$$F_0(\bx)=0.$$ \\
\FOR $j=1$ \TO $M$ \DO
\BEGIN
$$(\widehat{c}_j, \widehat{\theta}_j)= \underset{(c,\theta)}{\operatorname{argmin}}\sum_{i \in S_j(\eta)} L(y_i, F_{j-1}(\bx_i)+cf(\bx_i, \theta))$$ \\
$$T_j(\bx)=f(\bx, \widehat{\theta}_j)$$ \\
$$F_j(\bx)=F_{j-1}(\bx)+\nu\widehat{c}_j T_j(\bx)$$ \\
\END \\
\RETURN{$$\{T_j(\bx)\}_{j=1}^M$ and $F_M(\bx).$$}
\end{pseudocode}

Here $L(.,.)$ is a loss function, $S_j(\eta)$ is a subset of the indices $\{1,2,\ldots, n\}$ chosen by a sampling scheme $\eta,$ $0\leq \nu \leq 1$ is a memory parameter. 

The classic ensemble methods of Bagging, Random Forest, AdaBoost, and Gradient Boosting are special cases of ISLE ensemble model generation procedure \cite{seni2010ensemble}. In Bagging and Random Forests the weights in \ref{eq:additivemodel} are set to predetermined values, i.e. $w_0=0$ and $w_j=\frac{1}{M}$ for $j=1,2,\ldots,M.$ Boosting calculates these weights in a sequential fashion at each step by having positive memory $\nu,$ estimating $c_j$ and takes $F_M(\bx)$ as the final prediction model.  

Friedman \& Popescu \cite{friedman2003importance} recommend learning the weights $\{w_j\}_{j=0}^{M}$ using lasso \cite{tibshirani1996regression}. Let $T={\left(T_j(\bx_i) \right)_{i=1}^n}_{m=1}^M$ be the $n\times M$ matrix of predictions for the $n$ observations by the $M$ models in an ensemble. The weights $(w_0,\bw=\{w_m\}_{m=0}^{M})$ are obtained from \begin{equation}\label{modeltree}\hat{\bw}=\underset{\bw}{\operatorname{argmin}} (\by-w_0\bone_n-T\bw)'(\by-w_0\bone_n-T\bw)+\lambda \sum_{m=1}^M|w_m|.\end{equation}
$\lambda>0$ is the shrinkage operator, larger values of $\lambda$ decreases the number of models included in the final prediction model. The final ensemble model is given by \begin{equation}F(\bx)=w_0+\sum_{m=1}^{M}w_{m} T_m(\bx).\end{equation}

\subsection{Rule Ensembles}

The base learners in the preceding sections of this article can be used with any regression model, however usually they are used with regression trees. Each decision tree in the ensemble partitions the input space using the product of indicator functions of ''simple'' regions based on several input variables. A tree with $K$ terminal nodes define a $K$ partition of the input space where the membership to a specific node, say node $k,$ can be determined by applying the conjunctive rule $$r_k(\bx)=\prod_{l=1}^{p}I(x_l\in s_{lk}),$$ where $I(.)$ is the indicator function, $\bx=(x_1,x_2,\ldots, x_p)$ are the input variables. The regions $s_{lk}$ are intervals for a continuous variable and a subset of the possible values for a categorical variable. 

Given a set of decision trees, rules can be extracted from these trees to define a collection of rules. Let  $R={\left(r_k(\bx_i) \right)_{i=1}^n}_{k=1}^K$ be the $n\times K$ matrix of rules for the $n$ observations by the $K$ rules in the ensemble. The \bfseries rulefit \normalfont algorithm of Friedman \& Popescu \cite{friedman2008predictive} uses the weights $(w_0,\bw=\{ {w_k}\}_{k=0}^{K})$ that are estimated from \begin{equation}\label{modelrule}\hat{\bw}=\underset{\bw}{\operatorname{argmin}} (\by-w_0\bone_n-R\bw)'(\by-w_0\bone_n-R\bw)+\lambda \sum_{k=1}^K|w_k|\end{equation} in the final prediction model \begin{equation}F(\bx)=w_0+\sum_{k=1}^{K}w_{k} r_k(\bx).\end{equation}

\section{Post Processing Ensembles Revisited}

We can use the base learners in an ensemble as input variables in any regression method. Since the number of models in an ensemble can easily exceed the number of observations, we prefer regression methods that can handle high dimensional input.  A few such approaches like principal components, partial least squares regression, multivariate kernel smoothing and weighting are illustrated in this section. We will compare these approaches to the existing standards random forests and rulefit in the next section. 

\subsection{Principal Components and Partial Least Squares Regression}

The models in an ensemble are all aligned with the response variable and therefore we should expect that they are correlated with each other.  Principal component  regression (PCR) and partial least squares regression (PLSR) are two techniques which are suitable for high dimensional regression problems where the predictor variables exhibit multicollinearity. 

PCR and PLSR decompose the input matrix $X$ into orthogonal scores $T$ and loadings $P$  $$X = T P$$ and regress $Y$ on the first few columns of the loadings $P$ using ordinary least squares. This leads to biased but low variance estimates of the regression coefficients in model \ref{eq:additivemodel}. PLSR incorporates information on both $X$ and $Y$ in the loadings. 

Both of these methods behave as shrinkage methods \cite{friedman2001elements} where the amount of shrinkage is controlled by the number of loadings included.  An obvious question is to find the number of loadings needed to obtain the best generalization for the prediction of new observations. This is, in general, achieved by cross-validation techniques such as bootstrapping.

The illustrations following section demonstrate the good performance of PLSR for post processing trees or rules. PLSR, as opposed to lasso, achieves shrinkage without forcing sparsity on the input variables. The ensemble learners are all ''directed'' towards the output variable and therefore they exhibit strong multicollinearity. This is a case where we would expect PLSR to work better than lasso.

The coefficients of the tree ensemble model in \ref{modeltree} or the rule ensemble model in \ref{modelrule} can be used to evaluate importances of trees, rules and individual input variables \cite{friedman2008predictive}. For the tree ensembles the importance of the $k$th tree is evaluated as 
\[I_k=|w_k|std(T_k)\]measures the importance of the trees or rules, here $std(T_k)$ denotes the standard deviation for the output of the $k$th tree over the individuals in the training sample. For the rule ensembles the importance of the $k$th rule is calculated similarly as \[I_k=|w_k|\sqrt{s_k(1-s_k)}\] where $s_k=\frac{\sum_{i=1}^{n}{r_k(\bx_i)}}{n}$ is the support of rule $k.$ The individual variable importances are calculated from sum of the importances of the trees or rules which contain that variable. 

The PLSR model is in the same additive form as in \ref{eq:additivemodel}, therefore the weights $w_1, w_2, \ldots, w_M$ in the model can be used to calculate tree rule or variable importances the same way they were calculated for the lasso post processing approach.

\subsection{Multivariate Kernel Smoothing}

We will concentrate on kernel smoothing using the Nadaraya-Watson estimator.  For a detailed presentation of the subject, we refer the reader to (\cite{takezawa2006introduction}). The Nadaraya-Watson estimator is a weighted sum of the observed responses $\by.$ Let the value of base learners at an input point $\bx$ be written in a $M$ dimensional vector $t(\bx).$ The final prediction model at input point $\bx$ can be obtained as  $$F(\bx)=\frac{\sum_{i=1}^{n}K_h(t(\bx_i)-t(\bx))y_i}{\sum_{i=1}^{n}K_h(t(\bx_i)-t(\bx))}.$$ 

The kernel function $K_h(.)$ is a symmetric function that integrates to one, $h > 0$ is the smoothing parameter. In practice, the kernel function and the smoothing parameter are usually selected using the cross validated or bootstrap performances for a range of kernel functions and smoothing parameter values. 
 
\subsection{Weighting Ensembles using Out-of-Bag Observations}

As we have mentioned earlier, most of the earlier important ensemble methods combine the base models using weights. Both bagging and random forest algorithms use equal weighting.  Estimating $\hat{\bw}$ by minimizing $$\frac{1}{2}(\by-T\bw)'(\by-T\bw)$$ subject to the constraint $\bw\geq 0$ gives the Stacking approach of Wolpert \cite{wolpert1992stacked} and Breiman \cite{breiman1996stacked}. In stacking final prediction model is given by $$F(\bx)=T(\bx)\widehat{\bw}.$$ 

The ensemble generation algorithms based on bootstrapping the observations builds the base learners from the observations in the bootstrap sample, and leaves us with the out-of-bag observations to evaluate the generalization performance of that particular learner. The following weighting scheme will down weight the base learners which have bad generalization performance. Let $({y_{oob}}_i,{\bx_{oob}}_i)$ denote the $i$th out-of-bag observation for $i=1,2,\ldots, n_{oob}.$ We have $M$ base learners $\{T_l(\bx)\}_{l=1}^{M}.$ We can use $$F(\bx)=\frac{\sum_{l=1}^{M}\sum_{i=1}^{n_{oob}}K_h({y_{oob}}_i-T_l({\bx_{oob}}_i)) T_l(\bx)}{\sum_{l=1}^{M}\sum_{i=1}^{n_{oob}}K_h({y_{oob}}_i-T_l({\bx_{oob}}_i))}$$ as the prediction of the response at input value $\bx.$ This involves keeping track of the out-of-bag performance each model in the ensemble and using the weights $$w_l=\frac{\sum_{i=1}^{n_{oob}}K_h({y_{oob}}_i-T_l({\bx_{oob}}_i))}{\sum_{l=1}^{M}\sum_{i=1}^{n_{oob}}K_h({y_{oob}}_i-T_l({\bx_{oob}}_i))},$$ $l=1,2, \ldots, M.$

The value of $h$ controls the smoothness of the model. For large values of this parameter the kernel method will assign approximately equal weights to the learners $T_l,$ $l=1,2,\ldots, M$ and hence it is equivalent to random forest weighting. Smaller values of the parameter assigns higher weights to the models with small out of bag errors. It is customary to choose $h$ that minimizes the cross-validated or bootstrapped errors. In addition, it is sometimes beneficial to eliminate the models with lowest weights from the final ensemble.

\section{Illustrations}

The following ensemble models are compared in this section:
\begin{enumerate}
\item r(pslr): Partial Least Squares Regression with Rules,

\item t(pslr):  Partial Least Squares Regression with Trees,

\item r(lasso): lasso with Rules,

\item t(lasso):  lasso with Trees,

\item w(oob): Weighting Using Out-of-Bag performance,

\item wt(oob): Weighting Using Out-of-Bag performance (best $60\%$ of the trees),

\item rf: Random Forest,

\item ksr: Kernel Smoothing with Rules,

\item kst: Kernel Smoothing with Trees.
\end{enumerate}

In all these models hyper parameters of the models are set using 10 fold cross validation in the training sample. 

Our first example involves the Fusarium head blight (FHB) data set that is available from the author upon request. A very detailed explanation of this data set is given in \cite{janninkcap}. 

\begin{ex} FHB is a plant disease caused by the fungus Fusarium Graminearum and results in tremendous losses by reducing grain yield and quality. In addition to the decrease in grain yield and quality, another damage due to FHB is the contamination of the crop with mycotoxins. Therefore, breeding for improved FHB resistance is an important breeding goal. Our aim is to build a prediction model for FHB resistance in barley based on available genetic variables. The FHB data set included FHB measurements along with 2251 single nucleotide polymorphisms (SNP) on 622 elite North American barley lines.  The 10 fold cross validated accuracies measured by the correlations of true responses to the predicted values are displayed in Figure \ref{boxplot1}.

\begin{figure}[p]
	\centering
		\includegraphics[width=1.00\textwidth]{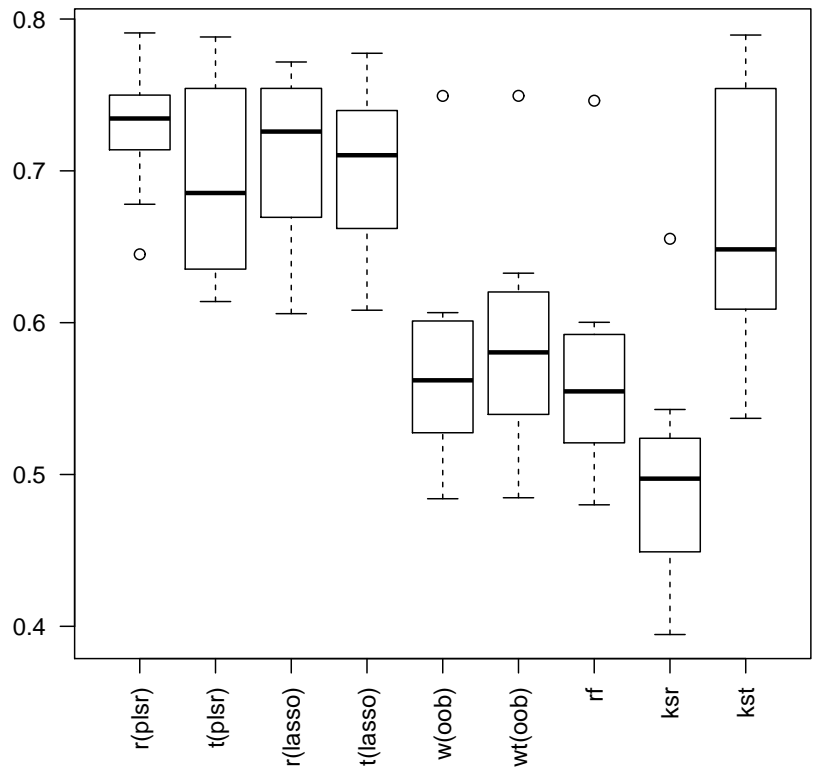}
		\caption{10 fold cross validated accuracies measured by correlation for the FHB data set. The ensemble of rules with PLSR has slightly higher accuracy compared to its alternative rules with lasso. The number of trees was set to $200.$ Maximum depth allowed for each tree or rule was set to 5.}
	\label{boxplot1}
\end{figure}
\end{ex}

\begin{ex}\label{ex2} In our second example we repeat the following experiment 100 times. Elements of the $150\times 100$ input matrix X are independently generated from  a $uniform(0,1)$ distribution. The elements of the coefficient matrix $\beta$ were also generated independently from $unif(0,1)$ and $85\%$ of these were selected randomly and set to zero. $150$ dimensional response vector $\by$ was generated according to $\by=X\beta+e$ where $e$ was generated from $N_{150}(\bzero,0.3I_{150})$ so that the signal ratio was about 2 to 1. The data was separated as training data and test data in the ratio of 2 to 1. The box plots in Figure \ref{ex2fig} compare the different approaches to ensemble post processing in terms of the accuracies in the test data set.  

\begin{figure}[p]
	\centering
		\includegraphics[width=1.00\textwidth]{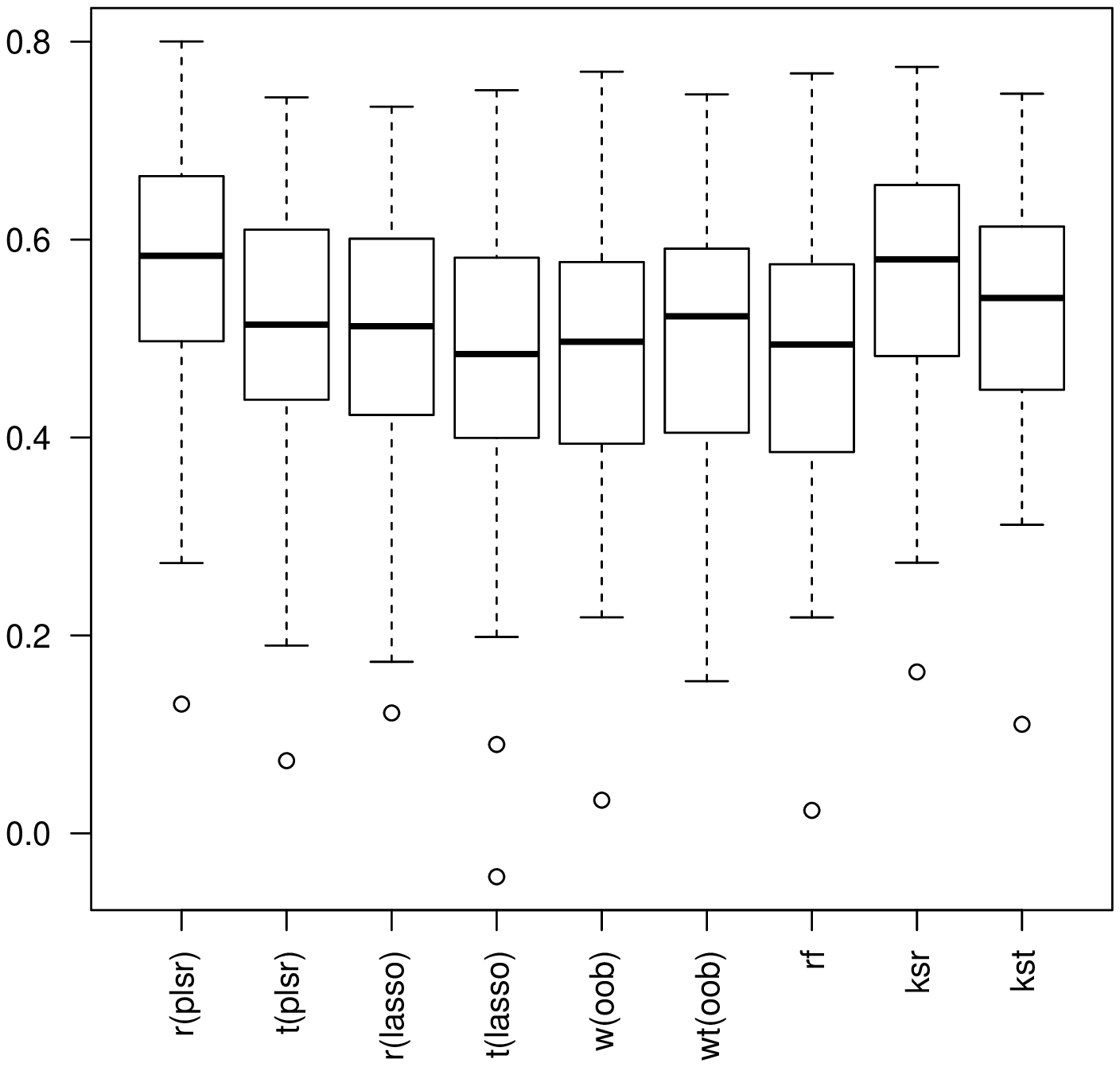}
	\caption{The box plots  in Figure \ref{ex2fig} compare the different approaches to ensemble post processing for the scenario in Example \ref{ex2}. The number of trees generated was 200, maximum depth parameter was set to 2.}
	\label{ex2fig}
\end{figure}

\end{ex}

\begin{ex}\label{ex3} In this example, we repeat the experiment in Friedman \& Popescu (\cite{friedman2008predictive}). Elements of the $1000\times 100$ input matrix are independently generated from $unif(0,1)$ distribution.  $1000$ dimensional response vector $\by$ was generated according to $\{y_i=10\prod_{j=1}^5 e^{-2x_{ij}^2}+\sum_{j=6}^{35}x_{ij}+e_i\}$ where $e_i$ was generated from $N(\bzero,\sigma^2=1).$ The data was separated as training data and test data in the ratio of 2 to 1. The box plots in Figure \ref{ex3fig} compare the test data performances of the different approaches over 100 replications of the experiment.  

\begin{figure}[p]
	\centering
		\includegraphics[width=1.00\textwidth]{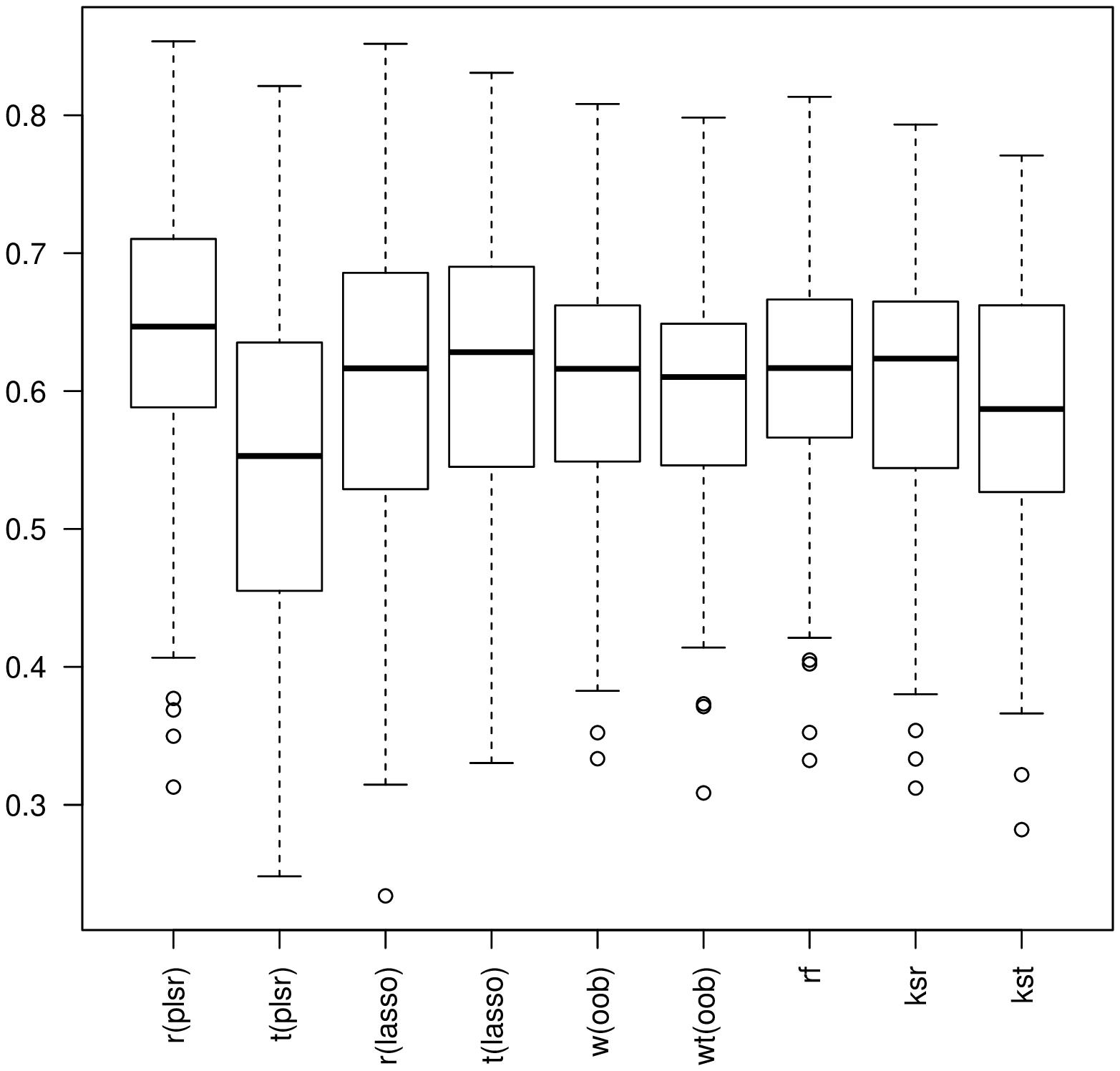}
	\caption{The box plots  in Figure \ref{ex2fig} compare the different approaches to ensemble post processing for the scenario in Example \ref{ex3}. Number of trees was set to 200, and the maximum depth parameter was set to 2.}
	\label{ex3fig}
\end{figure}

\end{ex}

\begin{ex} In order to compare the performance of prediction models we use the benchmark  data set ''Boston Housing'' (\cite{harrison1978hedonic}). This data set includes n=506 observations and p=14 variables. The response variable is the median house value from the rest of the 13 variables in the data set. 10 fold cross validated accuracies are displayed by the box plots in Figure \ref{boston}. The PLSR approach has the best cross validated prediction performance.

\begin{figure}[p]
	\centering
		\includegraphics[width=1.00\textwidth]{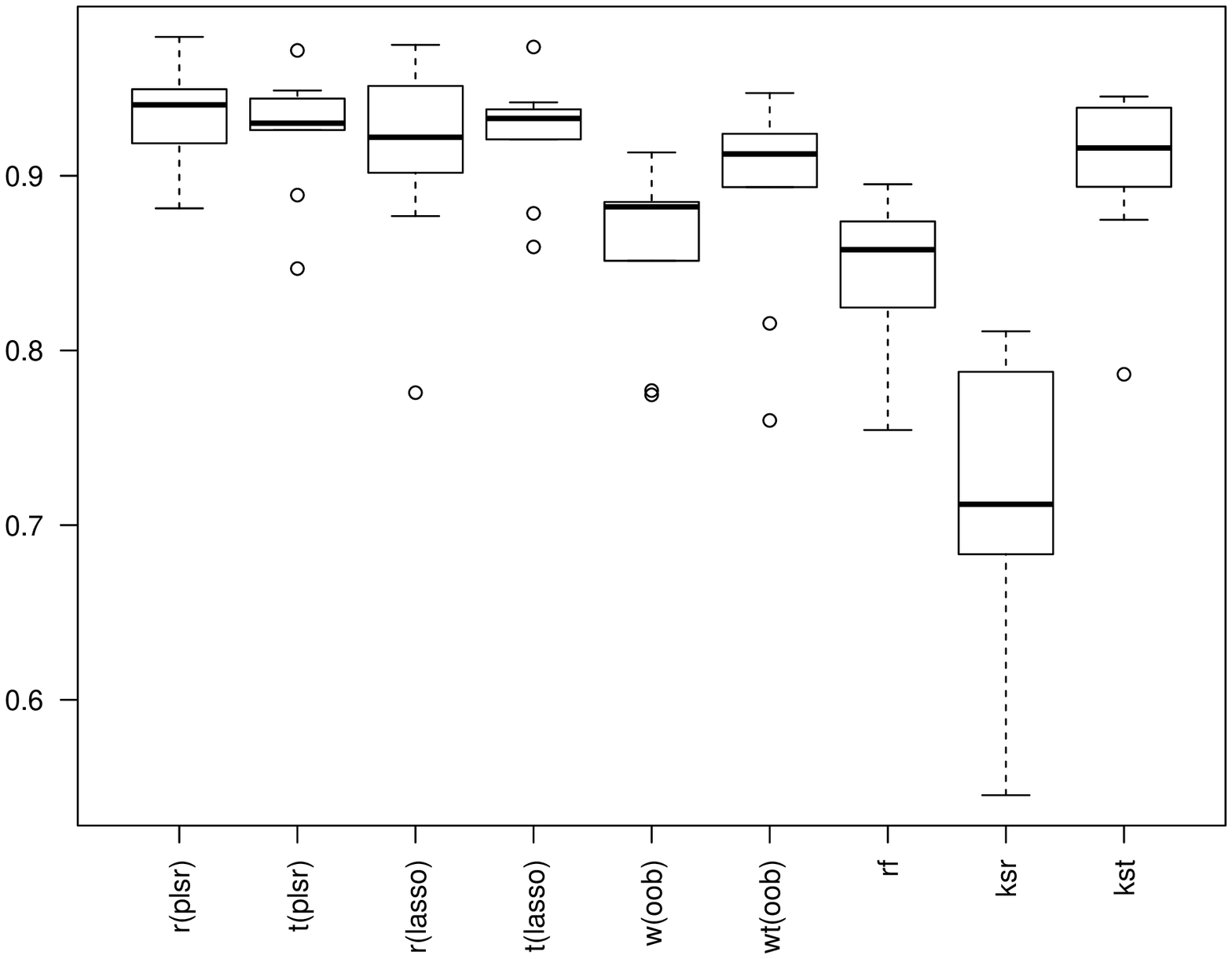}
	\caption{10 fold cross validated accuracies for the ''Boston Housing'' data are displayed by the box plots. The PLSR approach has the best cross validated prediction performance. We have generated 300 trees by the ISLE approach and maximum depth parameter was set to 4.  For the methods that use a kernel function we have uniformly used the Gaussian kernel. The sparsity parameters of the lasso or PLSR and the kernel width's parameters were obtained by minimizing 10 fold cross validated errors in the training data.}
	\label{boston}
\end{figure}

\end{ex}

\section{Conclusion} 
In this article, we have proposed several approaches for post processing a large ensemble of prediction models or rules. The approach taken here is to treat the ensemble models or the rules as base learners and use them as input variables in the regression problem. Some weighting approaches to ensemble models are also considered. 

The results from our simulations and benchmark experiments show that these post processing methods are promising. In most cases, the proposed models had better prediction performances than the ones given by the popular random forest or the rulefit algorithms. PLSR with rules uniformly produced the models with best prediction performances. The ensembles based on rules extracted from trees, in general, had better performances.  

The complexity of trees or rules in the ensemble increases with the increase in number of nodes from the root to the final node (depth). The maximum depth is an important parameter since it controls the degree of interactions between the input variables incorporated by the ensemble model and the its value should be set carefully.  It might also be useful to use some degree of cost pruning while generating the trees by the ISLE algorithm.

One last remark: This article argues that individual trees or rules should be treated as input variables to the statistical learning problem. It is almost always possible to incorporate other input variables like the original variables or their functions to our prediction model. The rulefit algorithm of Friedman \& Popescu optionally includes the input variables along with the rules in an additive model and uses lasso regression to estimate the coefficients in the model. Integrating additional input variables into the final ensemble is also straightforward with PLSR and kernel smoothing.

\section*{Acknowledgments}

I take this opportunity to express my gratitude to the people who have been instrumental in the successful completion of this article. This research was also supported by the USDA-NIFA-AFRI Triticeae Coordinated Agricultural Project, award number 2011-68002-30029. 

\bibliographystyle{plain}

\bibliography{ensemblebib}

\end{document}